\pdfoutput=1

\documentclass[11pt]{article}

\usepackage[]{acl}

\usepackage{times}
\usepackage{latexsym}
\usepackage{multirow}
\usepackage{booktabs}
\usepackage{tabularx}
\usepackage{graphicx}
\usepackage{makecell}
\usepackage{amsmath}
\usepackage{caption}
\usepackage[labelformat=simple]{subcaption}

\usepackage[T1]{fontenc}

\usepackage[utf8]{inputenc}

\usepackage{microtype}

\usepackage{inconsolata}

\title{Unleashing Large Language Models' Proficiency \\ in Zero-shot Essay Scoring} 

\author{Sanwoo Lee$^{1,2}$
Yida Cai$^{1,3}$
Desong Meng$^{1,2}$
Ziyang Wang$^{1,3}$
Yunfang Wu$^{1,2}$\thanks{~~Corresponding author.} \\
$^{1}$National Key Laboratory for Multimedia Information Processing, Peking University \\
$^{2}$School of Computer Science, Peking University \\
$^{3}$School of Software and Microelectronics, Peking University \\
\texttt{\{sanwoo, wuyf\}@pku.edu.cn},
\texttt{\{caiyida, 2100013162, wzy232303\}@stu.pku.edu.cn}
}

\begin{document}
\maketitle
\begin{abstract}
Advances in automated essay scoring (AES) have traditionally relied on labeled essays, requiring tremendous cost and expertise for their acquisition. Recently, large language models (LLMs) have achieved great success in various tasks, but their potential is less explored in AES. In this paper, we show that our zero-shot prompting framework, Multi Trait Specialization (MTS), elicits LLMs' ample potential for essay scoring. In particular, we automatically decompose writing proficiency into distinct traits and generate scoring criteria for each trait. Then, an LLM is prompted to extract trait scores from several conversational rounds, each round scoring one of the traits based on the scoring criteria. Finally, we derive the overall score via trait averaging and min-max scaling. Experimental results on two benchmark datasets demonstrate that MTS consistently outperforms straightforward prompting (Vanilla) in average QWK across all LLMs and datasets, with maximum gains of 0.437 on TOEFL11 and 0.355 on ASAP. Additionally, with the help of MTS, the small-sized Llama2-13b-chat substantially outperforms ChatGPT, facilitating an effective deployment in real applications.

\end{abstract}

\section{Introduction}

Automated essay scoring (AES) aims at evaluating and scoring essays with machine learning \citep{dikli2006overview}. AES is a promising alternative to costly and laborious human assessment, greatly resolving rater fatigue and inter-rater inconsistency. AES systems have been widely deployed in classroom settings \citep{dikli2014automated} and high-stakes tests such as TOEFL \citep{attali2006automated}.

\begin{figure}[!thbp]
    \centering
    \includegraphics[width=\columnwidth]{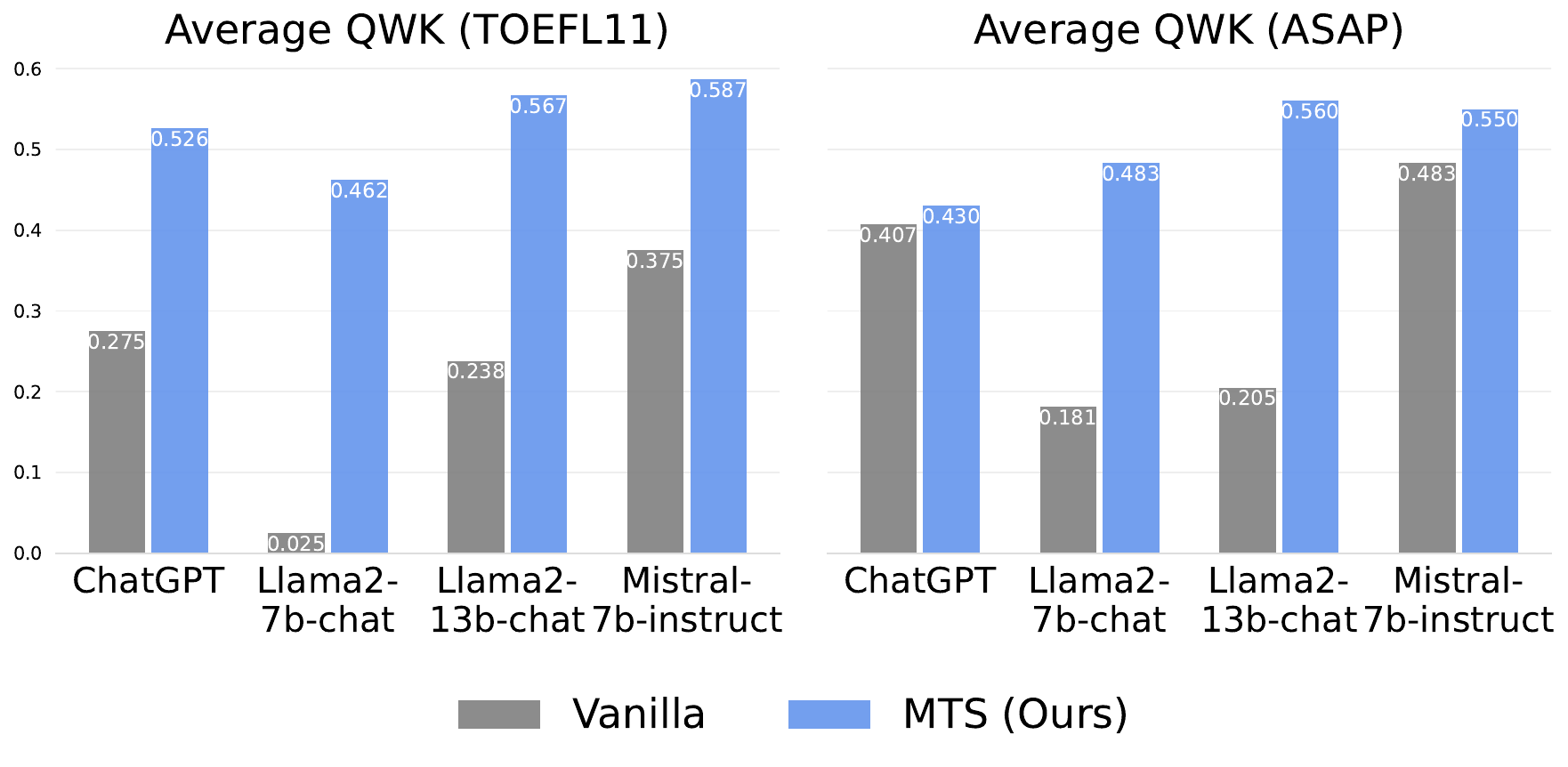}
    \caption{Comparison of our MTS zero-shot prompting framework and Vanilla baseline across different types of LLMs and datasets, measured on average QWK.}
    \label{fig:summary}
\end{figure}

Previous studies highly matched human ratings via developing supervised models tailored to a specific prompt \citep{yannakoudakis-etal-2011-new, taghipour-ng-2016-neural, dong-etal-2017-attention}, assuming essays from the train and test sets belong to the same prompt. However, prompt-specific models struggled when confronted with essays written for unseen prompts \citep{jin-etal-2018-tdnn, cozma-etal-2018-automated}. Hence AES is progressing towards reflecting more real-world scenarios, exemplified by cross-prompt approaches \citep{cao2020domain, jiang-etal-2023-improving} which reinforce domain transferability of the supervised models. 

Recently, advances in large language models (LLMs) \citep{brown2020language, ouyang2022training, openai2023gpt4, touvron2023llama} have led to a paradigm shift in which LLMs excel across a wide range of downstream tasks via zero-shot or few-shot instructions \citep{yuan-etal-2023-zero, zhang2023benchmarking}. In many cases, careful prompt design plays a crucial role in unlocking LLMs' potential. For instance, chain-of-thought (CoT) prompting \citep{wei2022chain, kojima2022large} improves LLMs' performance on complex reasoning benchmarks by externalizing the reasoning process. 

The development of LLM-based chatbots aligned with human preferences \citep{ouyang2022training, rafailov2023direct} has given rise to zero-shot AES, allowing us to move beyond the cross-prompt setting. However, leveraging LLMs for zero-shot AES is less explored, in contrast to proliferating studies harnessing LLMs to serve as an evaluation metric for machine-generated text \citep{chiang-lee-2023-large, zheng2023judging, liu-etal-2023-g}. Initial works of zero-shot AES \citep{mizumoto2023exploring, yancey2023rating} prompt LLMs to assign the overall score within a single step, which demonstrates suboptimal agreement with human raters. 

In this paper, we present \textbf{MTS} (\textbf{M}ulti \textbf{T}rait \textbf{S}pecialization), a zero-shot prompting framework to elicit essay scoring capabilities in LLMs, inspired by supervised models that explicitly predict trait scores and improve the overall scoring \citep{ridley2021automated, kumar-etal-2022-many, do-etal-2023-prompt}. In particular, we exploit ChatGPT \citep{ChatGPT} to decompose the writing quality into multiple traits and generate scoring criteria for each trait. Next, an LLM engages in several rounds of conversation, each round evaluating with respect to one of the traits. During the conversation, the LLM is instructed to retrieve quotes to provide faithful evaluation on the essay, then assign a score based on the given scoring criteria. Finally, the overall score is derived by averaging and min-max scaling the trait scores, in combination with the outlier clipping mechanism. 

We evaluate MTS on ASAP \citep{asap-aes} and TOEFL11 \citep{blanchard2013toefl11} with different LLMs, including ChatGPT, Llama 2 \citep{touvron2023llama} and Mistral 7b \citep{jiang2023mistral}. We take the Vanilla approach as a primary baseline which asks LLMs to produce rationales followed by an overall score. As illustrated in Figure~\ref{fig:summary}, MTS consistently outperforms Vanilla in average Quadratic Weighted Kappa (QWK) across all combinations of LLMs and datasets, with maximum gains of 0.437 ($0.025 \rightarrow 0.462$) on TOEFL11 and 0.355 ($0.205 \rightarrow 0.560$) on ASAP. In addition, the small-sized Llama2-13b-chat substantially surpasses ChatGPT with the help of MTS, enabling a more effective deployment. 


In essence, our contributions are as follows:
\begin{itemize}
    \item Our framework is free from training as well as labeling essays, which can be readily applied to new essay prompts (domains).
    \item We automatically decompose the AES task into more specific subtasks with respect to diverse traits, thereby  significantly boosting the agreement with human raters.
    \item We utilize min-max scaling with outlier clipping, effectively addressing LLMs' scoring bias and contributing to robust performance.
    \item MTS achieves promising results, largely exceeding Vanilla, also outperforming ChatGPT with the small-sized Llama2-13b-chat.  
\end{itemize}

\begin{figure}[!thbp]
    \centering
    \includegraphics[width=\columnwidth]{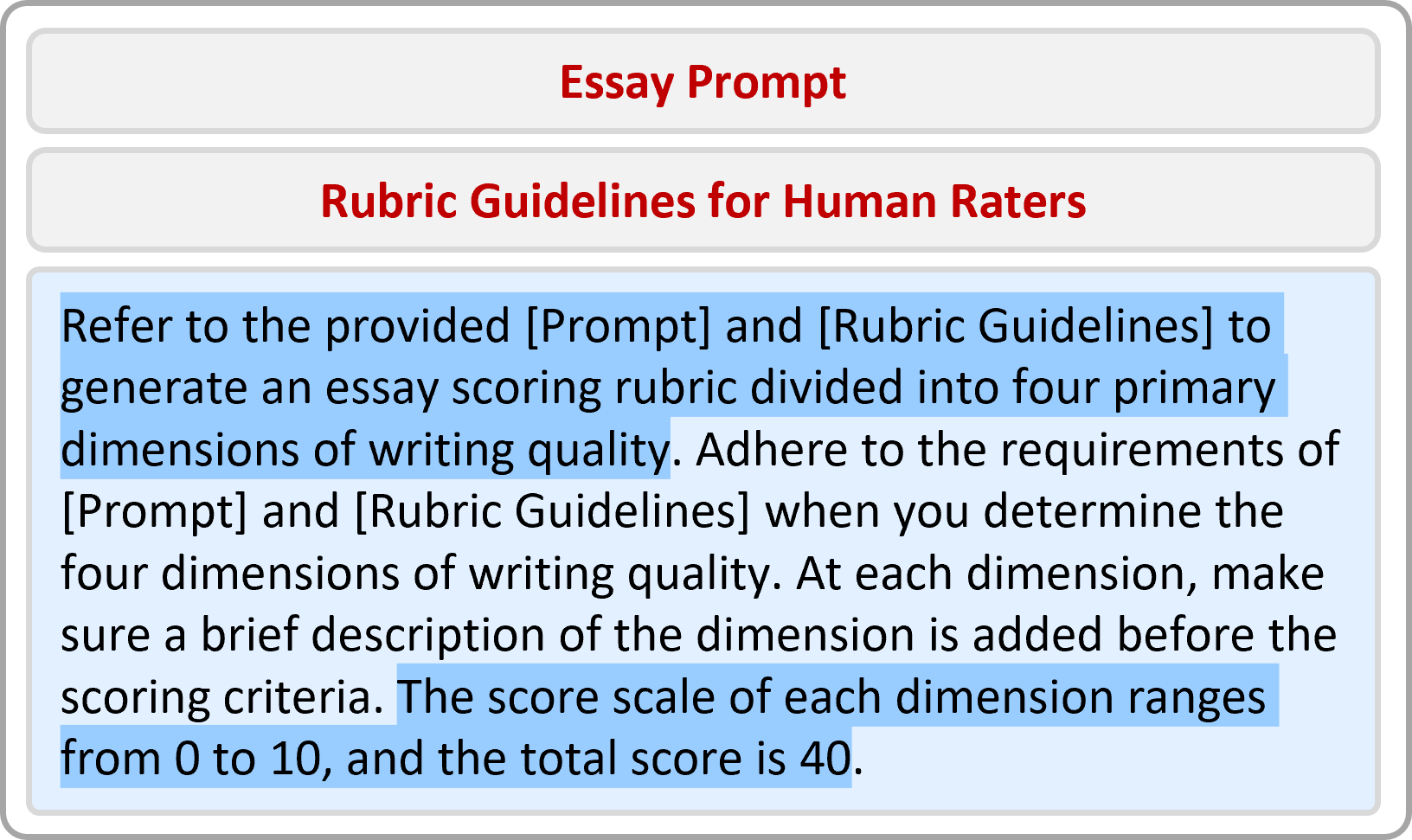}
    \caption{Illustration of the prompt for multi trait decomposition used for ASAP. The contents to be filled are denoted in \textcolor{red}{red}. See Appendix~\ref{appendix:mtd} for the templates used for ASAP and TOEFL11.}
    \label{fig:multi_trait_decomposition}
\end{figure}

\section{Method}\label{sec:method}

\begin{figure*}[htbp]
    \centering
    \includegraphics[width=\textwidth]{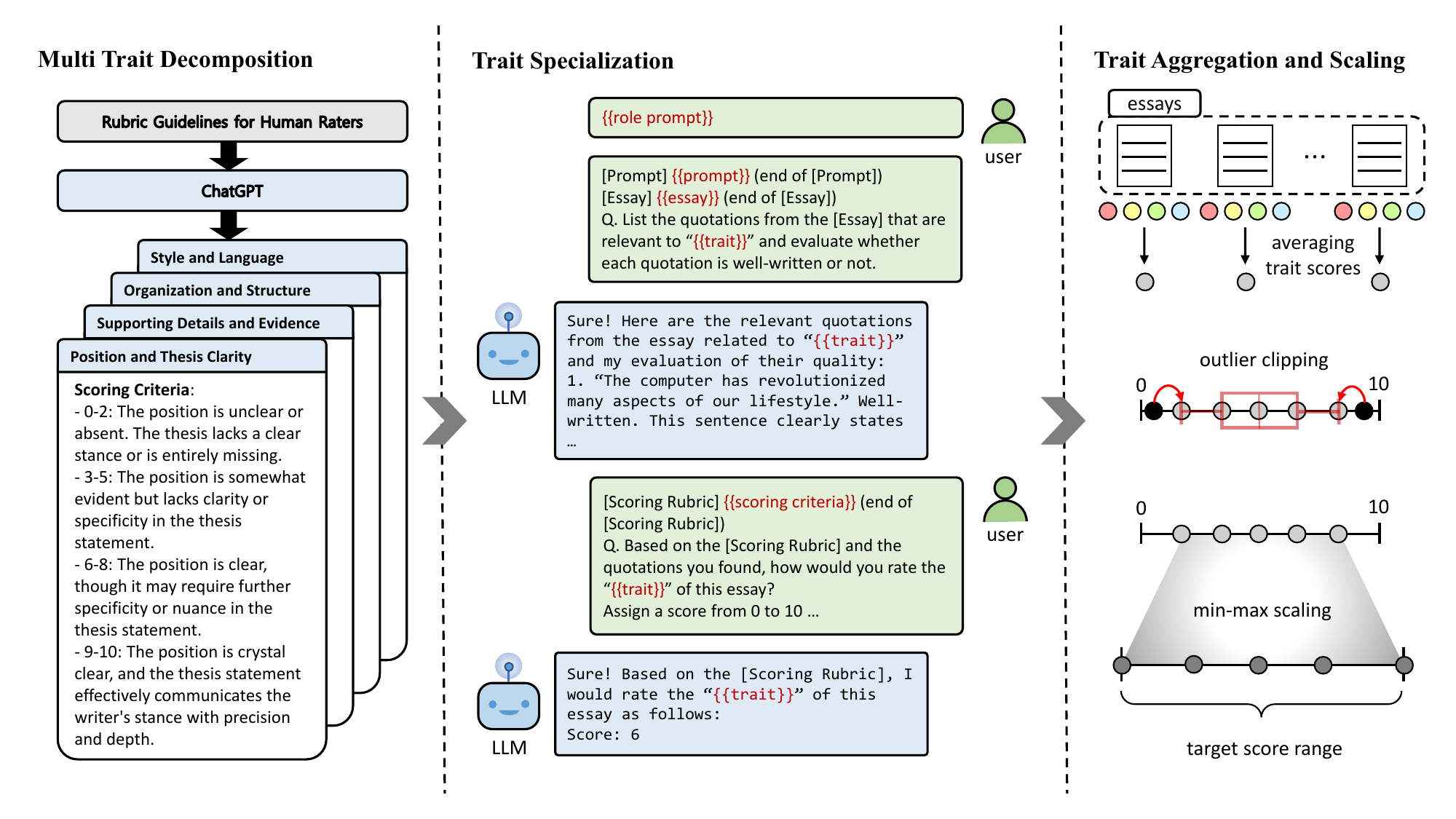}
    \caption{The illustration of Multi Trait Specialization framework. The parts to be filled with specific contents are substituted with comments between double curly braces and colored in red.}
    \label{fig:model}
\end{figure*}

We formalize the definition of zero-shot AES as follows: Given a dataset consisting of unlabelled essays $\mathcal{D}=\{x^{(i)}\}_{i=1}^{N_\mathcal{D}}$, the goal is to output an overall score $\hat{y}^{(i)}$ for every essay $x^{(i)}$ from a set of predefined scores $\mathcal{Y}$ where $\hat{y}^{(i)} \in \mathcal{Y}$. 

Multi trait specialization encourages LLMs to assess the essay from diverse aspects of writing quality. It consists of three steps: (1) decomposing writing proficiency into multiple traits and generating scoring criteria; (2) assigning a trait-specific score by step-by-step evaluation specialized to the trait; (3) deriving the final score via trait aggregation and scaling. The overall architecture of MTS is illustrated in Figure \ref{fig:model}.

\subsection{Multi Trait Decomposition} \label{sec:multi_trait_decomposition}

A straightforward way of zero-shot AES with LLM would be asking it to score an essay in a single response. Despite its simplicity, this approach cannot guarantee that the LLM employs the same scoring criteria across essays, leading to inconsistent evaluation. Moreover, scoring an essay should ideally be based on a comprehensive analysis of various dimensions of writing quality, while the LLM may be overloaded to do so in a single response.

We address this issue by decomposing the writing proficiency into several key traits and defining trait-specific scoring criteria which will be fixed during the assessment (details in Section \ref{sec:pvs}), contributing to a consistent scoring behavior of LLM. 
We automate this process via instructing ChatGPT to condense the rubric guidelines used by human raters into several key traits of writing quality and generate scoring criteria for each trait, using the prompt outlined in Figure~\ref{fig:multi_trait_decomposition}. This procedure ensures isolating trait-specific scoring criteria from rubric guidelines that mix multiple traits as a whole. An example of the generation result is depicted in the left part of Figure~\ref{fig:model} (see Appendix~\ref{appendix:guidance_mtd} for more results).



\subsection{Trait Specialization} \label{sec:pvs}
Prompt design plays a crucial role in unlocking the emergent abilities of LLMs. One of the key findings is that their reasoning ability benefits from subproblem decomposition of the complex promblem \citep{zhou2022least}. We hypothesize that LLMs provide more reasonable assessment of essays when specialized to perform step-by-step evaluation restricted to one aspect of writing quality, inspiring us to design the following steps of prompting (See Appendix~\ref{appendix:template_mts} for the full template):
\begin{enumerate}
    \item \textbf{Trait-specific Conversation}: For an essay, a number of independent conversations are initiated with each conversation specialized to one of the traits. Within the conversation, LLM is given a role prompt (i.e., system prompt) to focus solely on evaluating the specific trait.
    \item \textbf{Quote Retrieval}: Each conversation consists of two turns. In the first turn, LLM is required to retrieve quotes relevant to the trait and provide verbal evaluations for each quote.
    \item \textbf{Scoring}: In the second turn, LLM is asked to score the essay with respect to the trait, referring to both the previous turn and the given scoring criteria.
\end{enumerate}

In Step 1, leveraging trait-specific conversations not only simplifies the AES task as a form of subproblem decomposition but also prevents evaluations on each trait from being influenced by the other traits. In Step 2 and 3, the quote retrieval task is followed by the main scoring task, as depicted in the middle part of Figure~\ref{fig:model}. The quote retrieval task allows LLM to adhere to the details of the essay and avoid producing generic evaluation, whereas the scoring task transforms the verbal evaluation into the score based on the predefined scoring criteria.


\subsection{Trait Aggregation and Scaling}\label{sec:trait_aggregation_and_scaling}
Parsing the output of Section \ref{sec:pvs} yields trait scores $\{\hat{y}_{j}^{(i)}\}_{j=1}^{N_T}$ for an essay $x^{(i)}$, where $N_T$ represents the number of predefined traits. The trait scores should be transformed to an overall score $\hat{y}^{(i)}$ that falls under the target score range which may vary across different prompts (as in Table~\ref{tab:dataset_statistics}).

Based on the trait scores, we devise a simple yet effective trait aggregation and scaling strategy. First, the trait scores are aggregated by taking their average, and the outliers among the averaged scores are clipped using $\text{Q1}$ and $\text{Q3}$, i.e. the first and the third quartiles of the averaged scores, that is, the clipped score $\hat{y}_{agg}^{(i)}$ is computed as follows:  
\begin{equation}
    \hat{y}_{agg}^{(i)} = \min( \max(\frac{1}{N_T}\sum_{j=1}^{N_T} \hat{y}_j^{(i)}, v_{min}), v_{max})
\end{equation}
where $v_{min}=\text{Q1}-1.5(\text{Q3}-\text{Q1})$ and $v_{max}=\text{Q3}+1.5(\text{Q3}-\text{Q1})$. The value $1.5$ here is commonly used for outlier detection \citep{seo2006review}.
Next, the clipped scores are mapped to the target range $[a,b]$ via min-max scaling:
\begin{equation} \label{eq:2}
    \hat{y}^{(i)} = a + \frac{(\hat{y}_{agg}^{(i)} - \hat{y}_{min})(b-a)}{\hat{y}_{max} - \hat{y}_{min}}
\end{equation}
where $\hat{y}_{min}=\min\limits_{i} \hat{y}_{agg}^{(i)}$ and $\hat{y}_{max}=\max\limits_{i} \hat{y}_{agg}^{(i)}$. 
In this way, the clipping alleviates the sensitivity of min-max scaling to the outliers.

\section{Experimental Settings}
\subsection{LLMs}
To verify the effectiveness of our proposed method is universal across LLMs, we choose different types of LLMs which are not variants of one another: \textbf{ChatGPT}, \textbf{Llama 2} and \textbf{Mistral 7b}. In detail, we use their instruction-tuned models which are \textbf{gpt-3.5-turbo-0613}, \textbf{Llama2-7b-chat}, \textbf{Llama2-13b-chat} and \textbf{Mistral-7B-Instruct-v0.2}. The temperature is set to $0.1$ for all LLMs and the repetition penalty is set to $1.1$ for all LLMs but ChatGPT. Other hyperparameters for sampling follow the defaults. Experiments are run once with a fixed random seed due to limits of computational resources.

\subsection{Datasets and Evaluation Metric} \label{sec:dataset}
We conduct experiments on two datasets, ASAP\footnote{\url{https://www.kaggle.com/c/asap-aes/data}} and TOEFL11. ASAP (Automated Student Assessment Prize) comprises 12,978 essays provided by students spanning in grade level from 7 to 10. The essays were written in response to 8 prompts of varying genres and score ranges. TOEFL11 consists of 12,100 essays written by test takers of TOEFL iBT which measures the academic English proficiency of non-native English speakers. The statistics of the two datasets are shown in Table~\ref{tab:dataset_statistics}.

For ASAP, 10\% of the essays from each prompt are randomly sampled for testing. There is no significant difference in the average essay score between the sample and the population in all prompts (Z-test, $\alpha=0.05$). 
For TOEFL11 dataset, we adopt the test split from the 2013 Native Language Identification Shared Task \citep{tetreault2013report}, which consists of 1,100 essays collected from 8 prompts. 

We use Quadratic Weighted Kappa (QWK) to measure the agreement between groundtruth scores and predicted scores. QWK is commonly used in AES research \citep{taghipour-ng-2016-neural, dong-etal-2017-attention, cao2020domain,  xie-etal-2022-automated}. 

\begin{table}[t]
    \centering
    \small
    \resizebox{\columnwidth}{!}{
    \begin{tabular}{cccccc}
        \toprule
        \textbf{Dataset} & \textbf{Prompt} & \textbf{\#Essay} & \textbf{Genre} & \textbf{Avg Len} & \textbf{Range} \\
        \midrule
        \multirow{8}{*}{\textbf{ASAP}}
         & 1 & 1783 & ARG & 427 & 2-12 \\
         & 2 & 1800 & ARG & 432 & 1-6 \\
         & 3 & 1726 & RES & 124 & 0-3 \\
         & 4 & 1772 & RES & 106 & 0-3 \\
         & 5 & 1805 & RES & 142 & 0-4 \\
         & 6 & 1800 & RES & 173 & 0-4 \\
         & 7 & 1569 & NAR & 206 & 0-30 \\
         & 8 & 723 & NAR & 725 & 0-60 \\
        \midrule
        \multirow{8}{*}{\textbf{TOEFL11}}
         & 1 & 1656 & ARG & 342 & l/m/h \\
         & 2 & 1562 & ARG & 361 & l/m/h \\
         & 3 & 1396 & ARG & 346 & l/m/h \\
         & 4 & 1509 & ARG & 340 & l/m/h \\
         & 5 & 1648 & ARG & 361 & l/m/h \\
         & 6 & 960 & ARG & 360 & l/m/h \\
         & 7 & 1686 & ARG & 339 & l/m/h \\
         & 8 & 1683 & ARG & 344 & l/m/h \\
        \bottomrule
    \end{tabular}}
    \caption{Statistics of ASAP and TOEFL11. \textbf{Genre}: ARG (argumentative), RES (source-dependent), NAR (narrative). \textbf{Avg Len}: Average essay length in words. \textbf{Range}: Score range (l/m/h for low/medium/high).}
    \label{tab:dataset_statistics}
\end{table}

\begin{table*}[t]
    \centering
    \small
    \resizebox{\textwidth}{!}{
    \begin{tabular}{cccccccccccc}
        \toprule
        \textbf{Dataset} & \textbf{LLM} & \textbf{Method} & \textbf{P1} & \textbf{P2} & \textbf{P3} & \textbf{P4} & \textbf{P5} & \textbf{P6} & \textbf{P7} & \textbf{P8} & \textbf{Avg.}\\
        \midrule
        \multirow{9.5}{*}{\textbf{ASAP}}
         & \multirow{2}{*}{ChatGPT} 
         & Vanilla & 0.032 & 0.220 & 0.476 & 0.597 & 0.479 & 0.637 & \textbf{0.289} & \textbf{0.527} & 0.407 \\
         &  & MTS & \textbf{0.138} & \textbf{0.443} & \textbf{0.502} & \textbf{0.611} & \textbf{0.662} & \textbf{0.668} & 0.261 & 0.157 & \textbf{0.430} \\
         \cmidrule{2-12}
         & \multirow{2}{*}{Llama2-7b-chat} 
         & Vanilla & 0.163 & \textbf{0.468} & 0.016 & 0.000 & 0.117 & 0.304 & 0.187 & 0.192 & 0.181 \\
         &  & MTS & \textbf{0.371} & 0.466 & \textbf{0.504} & \textbf{0.378} & \textbf{0.673} & \textbf{0.507} & \textbf{0.563} & \textbf{0.409} & \textbf{0.483} \\
         \cmidrule{2-12}
         & \multirow{2}{*}{Llama2-13b-chat} 
         & Vanilla & 0.158 & 0.189 & 0.069 & 0.004 & 0.280 & 0.393 & 0.333 & 0.213 & 0.205 \\
         &  & MTS & \textbf{0.591} & \textbf{0.541} & \textbf{0.552} & \textbf{0.591} & \textbf{0.620} & \textbf{0.590} & \textbf{0.483} & \textbf{0.511} & \textbf{0.560} \\
         \cmidrule{2-12}
         & \multirow{2}{*}{Mistral-7b-instruct} 
         & Vanilla & 0.206 & \textbf{0.512} & 0.516 & 0.587 & 0.457 & 0.601 & 0.624 & \textbf{0.304} & 0.483 \\
         &  & MTS & \textbf{0.545} & 0.455 & \textbf{0.550} & \textbf{0.691} & \textbf{0.540} & \textbf{0.657} & \textbf{0.672} & 0.289 & \textbf{0.550} \\
         \midrule
         \midrule
         \multirow{13.5}{*}{\textbf{TOEFL11}}
         & \multirow{3}{*}{ChatGPT}
         & \citet{mizumoto2023exploring} & 0.308 & 0.165 & 0.252 & 0.182 & 0.181 & 0.336 & 0.318 & 0.318 & 0.258 \\
         &  & Vanilla & 0.215 & 0.240 & 0.337 & 0.332 & 0.227 & 0.306 & 0.237 & 0.306 & 0.275 \\
         &  & MTS & \textbf{0.495} & \textbf{0.447} & \textbf{0.651} & \textbf{0.595} & \textbf{0.489} & \textbf{0.496} & \textbf{0.500} & \textbf{0.536} & \textbf{0.526} \\
         \cmidrule{2-12}
         & \multirow{3}{*}{Llama2-7b-chat} 
         & \citet{mizumoto2023exploring} & 0.009 & 0.047 & 0.085 & 0.140 & 0.133 & 0.027 & 0.032 & 0.023 & 0.062 \\
         &  & Vanilla & 0.000 & -0.007 & 0.026 & -0.006 & 0.041 & 0.015 & 0.111 & 0.020 & 0.025 \\
         &  & MTS & \textbf{0.545} & \textbf{0.395} & \textbf{0.540} & \textbf{0.472} & \textbf{0.497} & \textbf{0.388} & \textbf{0.419} & \textbf{0.437} & \textbf{0.462} \\
         \cmidrule{2-12}
         & \multirow{3}{*}{Llama2-13b-chat} 
         & \citet{mizumoto2023exploring} & 0.125 & 0.132 & 0.400 & 0.130 & 0.462 & 0.176 & 0.113 & 0.123 & 0.208 \\
         &  & Vanilla & 0.196 & 0.156 & 0.285 & 0.268 & 0.165 & 0.329 & 0.249 & 0.257 & 0.238 \\
         &  & MTS & \textbf{0.580} & \textbf{0.373} & \textbf{0.703} & \textbf{0.557} & \textbf{0.612} & \textbf{0.457} & \textbf{0.620} & \textbf{0.630} & \textbf{0.567} \\
         \cmidrule{2-12}
         & \multirow{3}{*}{Mistral-7b-instruct} 
         & \citet{mizumoto2023exploring} & 0.227 & 0.218 & 0.383 & 0.350 & 0.222 & 0.129 & 0.132 & 0.196 & 0.232 \\
         &  & Vanilla & 0.486 & 0.259 & 0.355 & 0.344 & 0.431 & 0.456 & 0.286 & 0.383 & 0.375 \\
         &  & MTS & \textbf{0.637} & \textbf{0.510} & \textbf{0.654} & \textbf{0.587} & \textbf{0.516} & \textbf{0.554} & \textbf{0.564} & \textbf{0.677} & \textbf{0.587} \\
         \bottomrule
    \end{tabular}}
    \caption{Zero-shot evaluation results in QWK. \textbf{P1-8} denotes Prompt 1-8. The best measures in each LLM are in bold. Negative value indicates that the predictions are worse than random.}
    \label{tab:main_results}
\end{table*}

\subsection{Implementation Details} \label{sec:implementation_details}
\noindent\textbf{Scoring Criteria.} As outlined in Section \ref{sec:multi_trait_decomposition}, ChatGPT generates scoring criteria based on the rubric guidelines for human raters. For ASAP, we use the dataset's original rubric guidelines for each prompt. For TOEFL11, we follow \citet{mizumoto2023exploring} and choose IELTS Task2 Writing Band Descriptor as the rubric guidelines instead of TOEFL Independent Writing Rubrics since the former provides more fine-grained descriptions. Note that each prompt in ASAP has a distinct scoring criteria whereas all prompts in TOEFL11 share the same scoring criteria.


\noindent\textbf{Scoring Strategy.}
For ASAP, the predicted score $\hat{y}$ is rounded to integers. For TOEFL11, $\hat{y}$ is first scaled to $[1, 5]$ \citep{blanchard2013toefl11}, then mapped to low/medium/high with respect to the thresholds of 2.25 and 3.75. 

\begin{figure}[!t]
    \centering
    \includegraphics[width=\columnwidth]{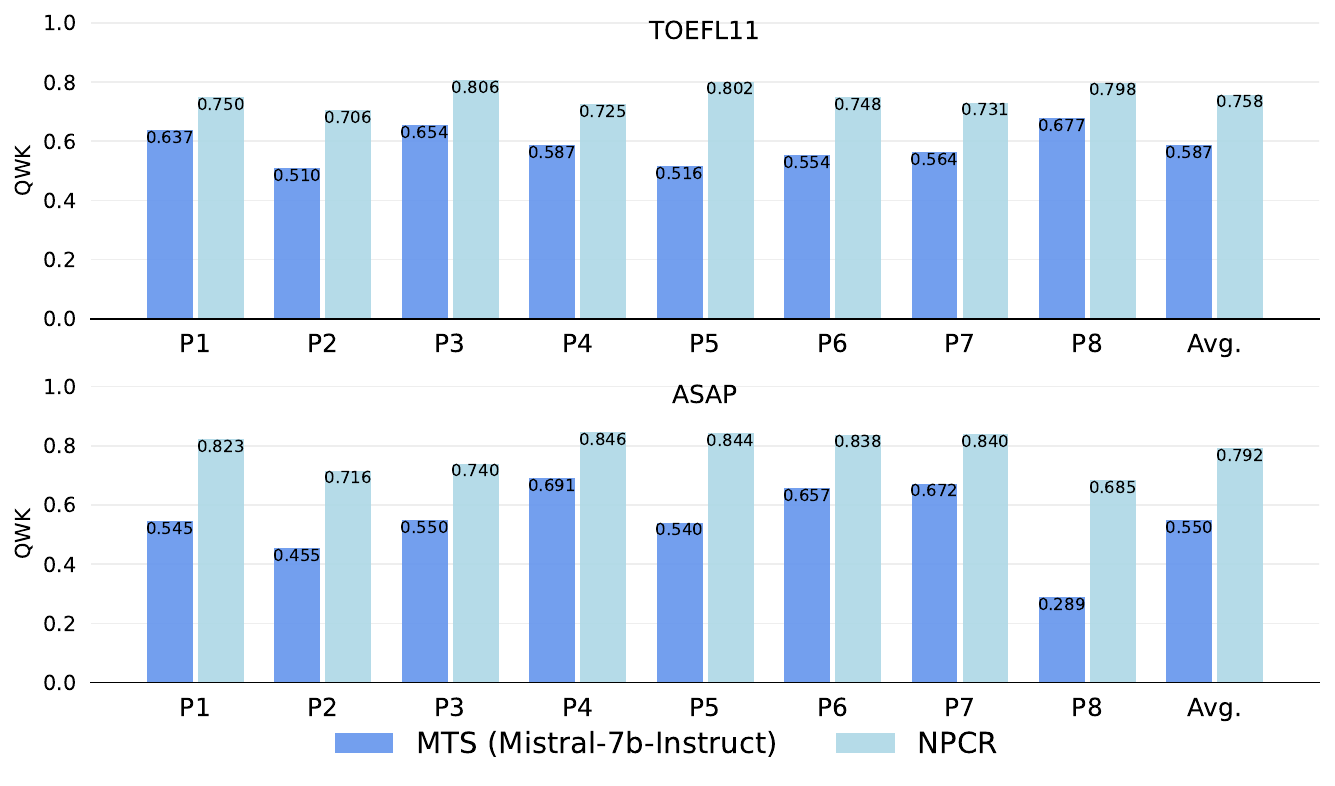}
    \caption{Comparison between MTS (Mistral-7b-instruct) and supervised SOTA (NPCR).}
    \label{fig:comparison_with_sota}
\end{figure}

\subsection{Comparisons with Other Methods}
We compare MTS against two LLM-based zero-shot baselines as well as a supervised SOTA model. 

\noindent\textbf{Rubric Scoring} \citep{mizumoto2023exploring} is a zero-shot approach designed for TOEFL11 which provides the LLM with human rubrics and asks it to assign an overall score within the target range. 

\noindent\textbf{Vanilla} is a zero-shot approach that asks the LLM to assign an overall score within the target range and provide rationales before the score to elicit CoT. Requiring rationales has been shown to improve zero-shot scoring performance on the CEFR scale \citep{yancey2023rating}. 
See Appendix~\ref{appendix:template_vanilla} for the full template.

\noindent\textbf{NPCR} \citep{xie-etal-2022-automated} is the state-of-the-art supervised prompt-specific model which predicts the score difference of two input essays based on pairwise ranking objective. For a valid comparison, we re-implement NPCR to ensure it is evaluated on the same test set as MTS. See Appendix~\ref{appendix:details_npcr} for the training details.


\section{Main Results}

The zero-shot evaluation results are shown in Table~\ref{tab:main_results}. Building upon asking the LLM to assign a score, providing human rubrics \citep{mizumoto2023exploring} and requiring rationales (Vanilla) yield similar average QWK on TOEFL11. MTS not only leverages the scoring rubrics but also retrieves evidence in trait-specific manner, consistently and significantly outperforming the baseline(s) in average QWK across all LLMs, on both datasets. In general, MTS achieves great performance gains over the baselines for Llama 2 series and moderate gains for ChatGPT and Mistral-7b-Instruct. For instance, MTS greatly improves over Vanilla using Llama2-13b-chat, with gains of 0.355 ($0.205\rightarrow0.560$) on ASAP and 0.329 ($0.238\rightarrow0.567$) on TOEFL11. This reassures the importance of careful prompt design to elicit LLM's ample potential to perform AES. Moreover, the superiority of MTS remains solid across diverse settings of AES reflected in the datasets, including variations in essay genre and the first language (L1) backgrounds of the test takers. 

By comparing different LLMs, we observe that Mistral-7b-instruct achieves the most competitive performance under different prompting methods overall, reaching average QWK of 0.550 on ASAP and 0.587 on TOEFL11. Interestingly, ChatGPT underperforms the small-sized Mistral-7b-Instruct across all methods but Rubric Scoring, implying that the model size might not be a decisive factor in the performance. Nevertheless, the larger model tends to perform better within the same model type, evidenced by the comparison between Llama2-7b-chat and Llama2-13b-chat. 

In terms of the comparison with the supervised SOTA, the gap in average QWK between MTS and NPCR can be narrowed down to 0.171 on TOEFL11 and 0.242 on ASAP using Mistral-7b-instruct, as shown in Figure~\ref{fig:comparison_with_sota}. This can be promising given that NPCR consumes approximately a thousand labeled essays per prompt to reach the SOTA. We expect further reduction in the gap with more powerful LLMs.

\begin{table*}[t]
    \centering
    \small
    \resizebox{\textwidth}{!}{
    \begin{tabular}{lccccccccc}
        \toprule
        \textbf{Method} & \textbf{P1} & \textbf{P2} & \textbf{P3} & \textbf{P4} & \textbf{P5} & \textbf{P6} & \textbf{P7} & \textbf{P8} & \textbf{Avg.}\\
        \midrule
         Vanilla & 0.158 & 0.189 & 0.069 & 0.004 & 0.280 & 0.393 & 0.333 & 0.213 & 0.205 \\
         Scoring All Traits Sequentially & 0.267 & 0.237 & 0.354 & 0.280 & 0.292 & 0.396 & \textbf{0.512} & 0.111 & 0.306\\
         Scoring Each Trait Independently & 0.489 & 0.487 & 0.478 & 0.467 & 0.568 & 0.549 & 0.471 & 0.430 & 0.492 \\
         \hspace{3mm} +Quote Retrieval and Scoring & \textbf{0.591} & \textbf{0.541} & \textbf{0.552} & \textbf{0.591} & \textbf{0.620} & \textbf{0.590} & 0.483 & \textbf{0.511} & \textbf{0.560} \\
         \bottomrule
    \end{tabular}}
    \caption{Ablation study of Trait Specialization. The QWKs of Llama2-13b-chat measured on ASAP are reported. The best performance in each column is in bold. See Appendix \ref{appendix:template_sequential} and \ref{appendix:template_independent} for the templates of 2nd and 3rd method.}
    \label{tab:effect_of_design_choices}
\end{table*}

\section{Analysis}
We conduct experiments to derive insights into the success of MTS by analyzing each of the modules in MTS. 

\subsection{Analysis of Multi Trait Decomposition}

MTS requires LLMs to follow the predefined traits and scoring criteria (i.e., the \textbf{guidance}), the quality of which is important to elicit their potential. 
In this section, we examine the effect of different guidance generation methods.

\begin{figure}[!thbp]
    \centering
    \includegraphics[width=\columnwidth]{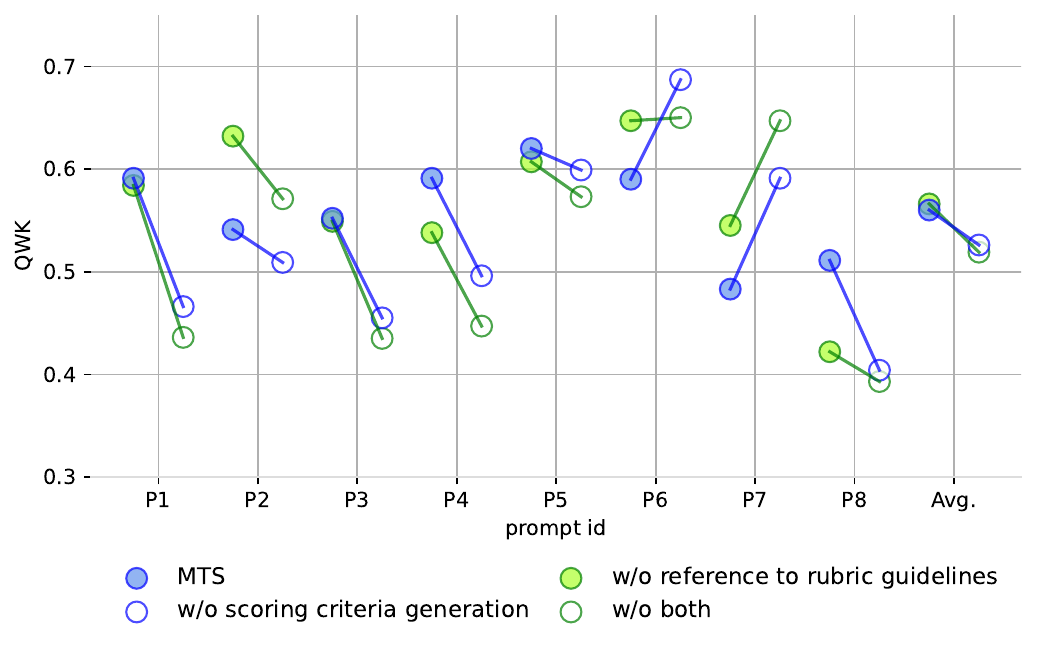}
    \caption{Ablation over (1) reference to
    rubric guidelines and (2) scoring criteria generation. The QWKs of Llama2-13b-chat on ASAP are reported.}
    \label{fig:ref_cri}
\end{figure}
 
\paragraph{Role of Rubric Guideline and Scoring Criteria.}
When decomposing the writing proficiency into multiple traits, MTS (1) refers to rubric guidelines designed by human raters, and (2) generates scoring criteria along with each trait. We conduct ablation study over both options to investigate impact on the performance. 

Figure~\ref{fig:ref_cri} shows that the reference to the rubric guidelines has negligible impact on the average QWK. Yet skipping the reference leads to higher standard deviation ($0.044 \rightarrow 0.066$) of the QWKs, indicating a fluctuating performance across the prompts. 
Next, the average QWK drops greatly after discarding the scoring criteria, both with and without the reference to human standards. One possible reason could be that the scoring criteria regulates the LLMs' behavior, encouraging them to adhere to the predefined criteria for better consistency. 

\begin{figure}[!thbp]
    \centering
    \begin{subfigure}{.47\columnwidth}
        \includegraphics[width=\columnwidth]{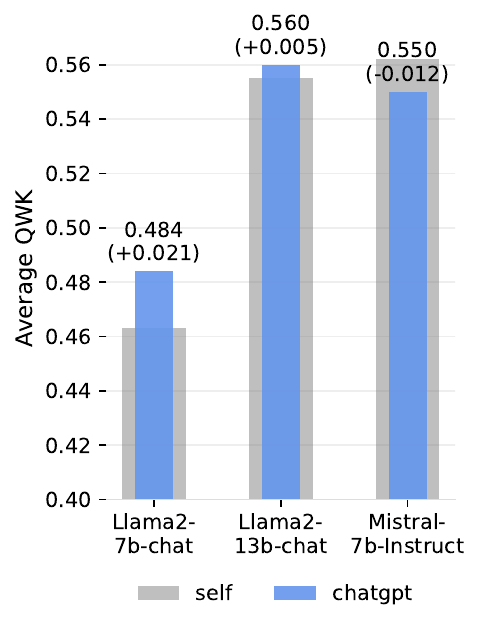}
        \caption{Impact of the source of the guidance on average QWK.}
        \label{fig:guidance_source}
    \end{subfigure}
    \hfill
    \begin{subfigure}{.47\columnwidth}
        \includegraphics[width=\columnwidth]{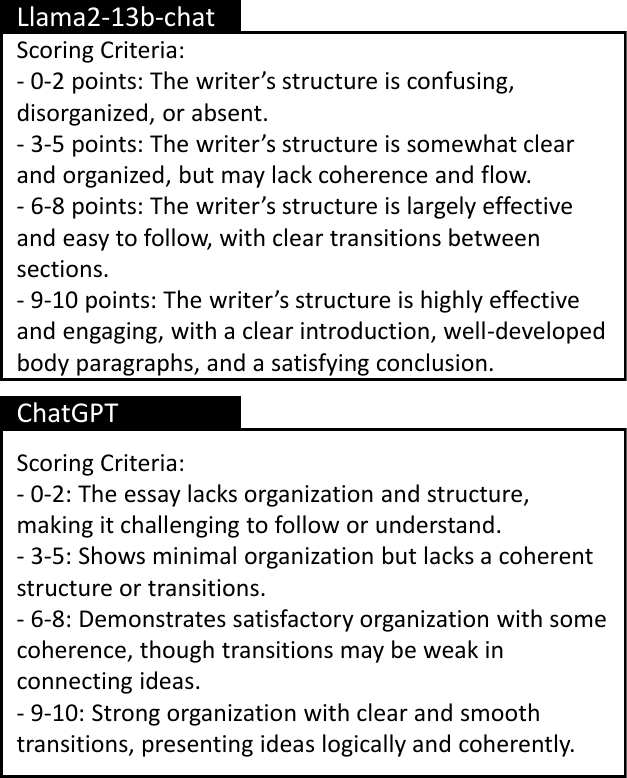}
        \caption{The scoring criteria for "Organization and Structure" generated by Llama2-13b-chat and ChatGPT, respectively.}
        \label{fig:guidance_example}
    \end{subfigure}
    \caption{Comparison of guidance (traits and scoring criteria) generated by ChatGPT and LLM itself.}
    \label{fig:self_versus_chatgpt}
\end{figure}

\paragraph{Leveraging ChatGPT for Guidance Generation.} 
MTS leverages ChatGPT to generate guidance 
for all LLMs used for actual scoring. We assess the significance of the guidance generated by ChatGPT over the one generated by the LLM itself. Figure \ref{fig:guidance_source} shows that
ChatGPT-generated guidance slightly outperforms self-generated ones for Llama2-7b-chat and Llama-13b-chat, whereas the opposite holds for Mistral-7b-Instruct. Figure \ref{fig:guidance_example} further reveals that both ChatGPT and Llama2-13b-chat produce reasonable scoring criteria. Therefore, while ChatGPT-generated guidance is readily applicable to various LLMs, it is still valid for the LLMs to use the self-generated guidance instead.

\subsection{Ablation Study of Trait Specialization} \label{sec:ablation_study}
With the scoring guidance given, the structure of Trait Specialization has evolved over Vanilla with a sequence of incremental improvements:

\noindent\textbf{Scoring All Traits Sequentially.} The LLM is required to read through the entire guidance and generate the evaluation-score pairs for all traits sequentially in a single turn of a conversation. Trait scores assigned in the fixed range of $[0, 10]$ are aggregated and scaled the same way as MTS.

\noindent\textbf{Scoring Each Trait Independently.} For each trait, a new trait-specific conversation is initiated where the LLM reads through the guidance restricted to the specific trait and generate the evaluation-score pair for the trait in a single turn. Trait scores are aggregated and scaled the same way as MTS.

\noindent\textbf{Quote Retrieval and Scoring.} On the basis of scoring each trait independently, we divide each conversation into \textbf{two turns} where the quote retrieval task precedes the scoring task, instead of generating evaluation-score pair in a single turn. This constitutes our proposed MTS.

Table~\ref{tab:effect_of_design_choices} demonstrates that all of the above design choices have positive impact on the average QWK. Building on Vanilla, we have considered two ways of integrating the predefined guidance into the conversation: scoring all traits sequentially and scoring each trait independently. While both methods are beneficial, independently scoring each trait proves to be much more effective, ensuring the evaluation on one trait is not influenced by the other traits. In addition, quote retrieval and scoring further enhances performance on all prompts, highlighting the importance of an in-depth analysis of the essay's content prior to assigning the score.

A characteristic these three strategies share in common is to decompose a complex problem of assigning an overall score into simpler subproblems and more specific tasks. We observe that this idea succeeds in zero-shot AES, in addition to complex reasoning tasks \citep{zhou2022least}. 

\begin{figure}[!thbp]
    \centering
    \includegraphics[width=\columnwidth]{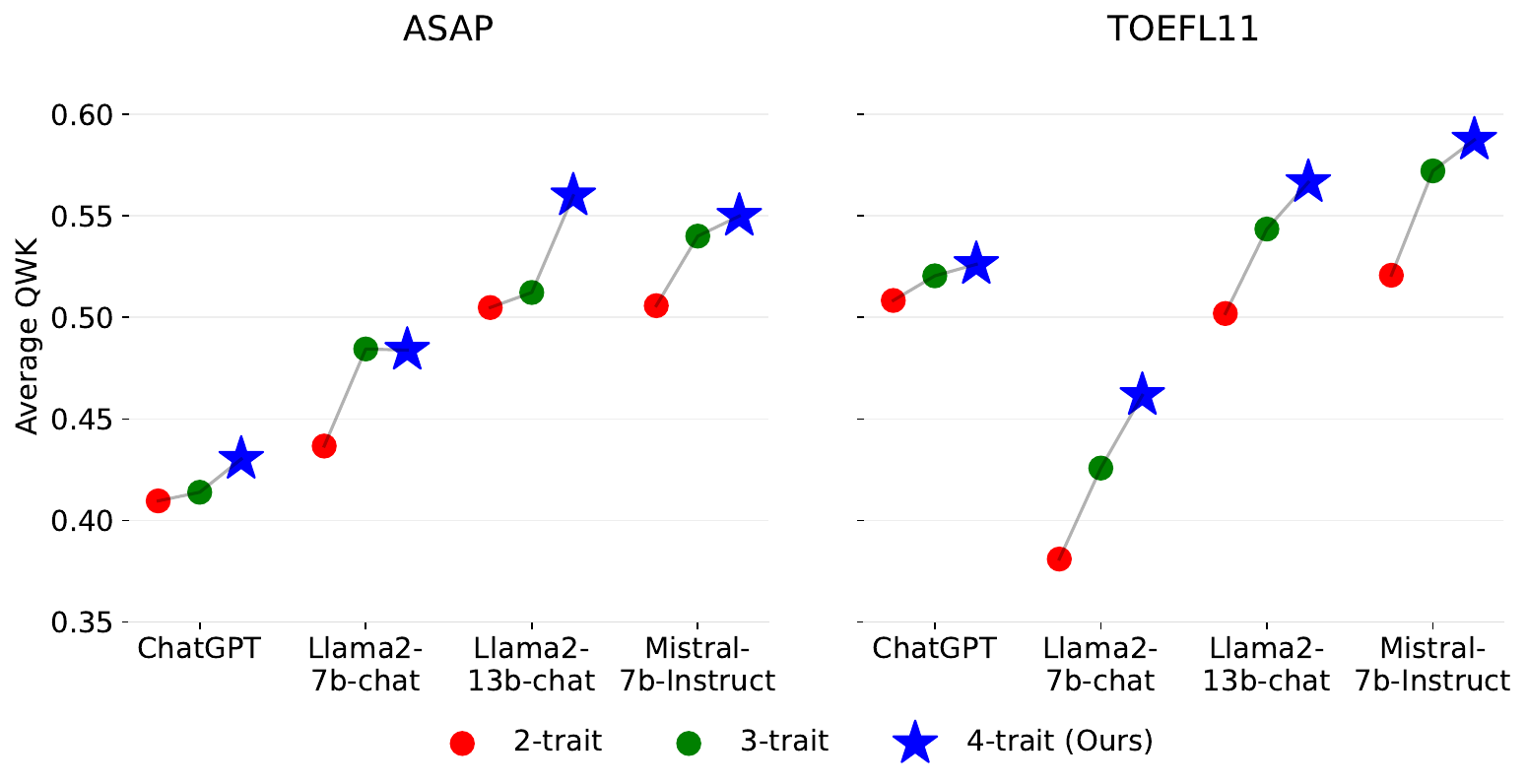}
    \caption{QWK under different numbers of traits, averaged over 8 prompts. $n$-trait denotes further averaging the performance over all combinations ($\binom{4}{n}$) of $n$ traits.}
    \label{fig:aggregation}
\end{figure}

\begin{table}[t]
    \centering
    \small
    \resizebox{\columnwidth}{!}{
\begin{tabular}{l|cccc|cccc}
\toprule
\multirow{2}{*}{}  & \multicolumn{4}{c|}{\textbf{ASAP}}                                                                    & \multicolumn{4}{c}{\textbf{TOEFL11}}                                                                 \\
\cmidrule{2-9}
                   & \textbf{C} & \textbf{L7} & \textbf{L13} & \textbf{M7} & \textbf{C} & \textbf{L7} & \textbf{L13} & \textbf{M7} \\
\cmidrule{1-9}
\textbf{Vanilla+}    & \textbf{0.455}            & 0.189                   & 0.303                    & 0.522                        & 0.463   & 0.164                   & 0.196                    & 0.496                        \\
\textbf{MTS} & 0.430   & \textbf{0.484}          & \textbf{0.560}           & \textbf{0.550}               & \textbf{0.526}   & \textbf{0.462}          & \textbf{0.567}           & \textbf{0.587}               \\
\bottomrule

\end{tabular}
    }
    \caption{ Average QWK of the overall (\textbf{Vanilla+}) and diversified (\textbf{MTS}) assessment. \textbf{C}: ChatGPT; \textbf{L7/13}: Llama2-7/13b-chat; \textbf{M7}: Mistral-7b-instruct. Best QWK in each column is in bold.}
    \label{tab:vanilla_versus_mts}
\end{table}

\subsection{Analysis of Trait Aggregation and Scaling}
\paragraph{Merit of Diversified Assessment.} MTS averages multiple trait scores $\{\hat{y}_{j}^{(i)}\}_{j=1}^{N_T}$ and scales the averaged scores to the target score range. We examine whether aggregating more trait scores improves performance by selecting all subsets of the original four traits with varying cardinality of $n\in\{2,3,4\}$ and evaluating the average QWK for each cardinality, as shown in Figure~\ref{fig:aggregation}. We observe a clear tendency where a higher number of traits leads to elevated performance, suggesting that different traits are complementary to each other. In other words, MTS takes advantages of diversified assessment of writing proficiency.

To further inspect if diversified scoring brings more benefit than the overall evaluation, we consider a new baseline called \textbf{Vanilla+} which predicts the \textit{overall} score in the same range as MTS (from 0 to 10) and applies the same scaling method as MTS\footnote{Since the overall scores given by Vanilla+ lack diversity, applying outlier clipping may result in all scores being clipped to a single value for some prompts, in which case we set their QWKs to zero. }. As shown in Table~\ref{tab:vanilla_versus_mts}, the average of trait scores leads to better estimates of the writing quality than the overall score in most cases.

\begin{table}[t]
    \centering
    \small
    \resizebox{\columnwidth}{!}{
\begin{tabular}{l|cccc|cccc}
\toprule
\multirow{2}{*}{}  & \multicolumn{4}{c|}{\textbf{ASAP}}                                                                    & \multicolumn{4}{c}{\textbf{TOEFL11}}                                                                 \\
\cmidrule{2-9}
                   & \textbf{C} & \textbf{L7} & \textbf{L13} & \textbf{M7} & \textbf{C} & \textbf{L7} & \textbf{L13} & \textbf{M7} \\
\cmidrule{1-9}
\textbf{fixed}     & 0.350            & 0.254                   & 0.477                    & 0.520                        & 0.357            & 0.071                   & 0.445                    & 0.385                        \\
\textbf{minmax}    & 0.405            & 0.477                   & 0.553                    & 0.529                        & \textbf{0.526}   & 0.438                   & 0.420                    & 0.499                        \\
\textbf{+clipping} & \textbf{0.430}   & \textbf{0.484}          & \textbf{0.560}           & \textbf{0.550}               & \textbf{0.526}   & \textbf{0.462}          & \textbf{0.567}           & \textbf{0.587}               \\
\bottomrule
\end{tabular}
    }
    \caption{Average QWK under different scaling methods. \textbf{clipping}: outlier clipping. Best QWK in each column is in bold.}
    \label{tab:scaling}
\end{table}

\begin{figure}[!t]
    \centering
    \includegraphics[width=\columnwidth]{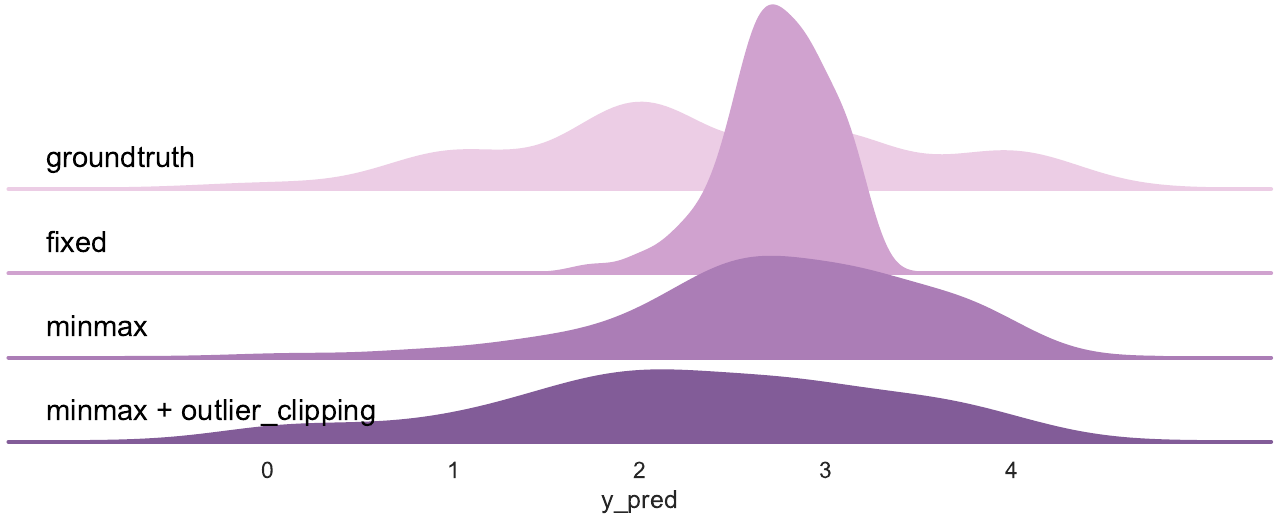}
    \caption{Distributions (KDE) of $\hat{y}$, estimated on ASAP Prompt 5 using Llama2-7b-chat.}
    \label{fig:scaling_distribution}
\end{figure}



\paragraph{Significance of Min-max Scaling and Outlier Clipping.} We conduct a comparative analysis of various scaling methods. We consider a simple baseline of \textbf{fixed scaling} where $\hat{y}_{min}$ and $\hat{y}_{max}$ are fixed to $0$ and $10$ during min-max scaling. As shown in Table~\ref{tab:scaling}, fixed scaling mostly fails, especially with Llama2-7b-chat showing significantly degraded performance. Conversely, min-max scaling greatly outperforms fixed scaling in average QWK for most cases, and clipping outliers further brings consistent and sometimes crucial improvements over min-max scaling.

Figure \ref{fig:scaling_distribution} shows Kernel Density Estimation (KDE) of $\hat{y}$, providing insights into how the scaling mechanism of MTS works: 
(1) The LLM displays a bias toward predicting within a specific and concentrated interval, as shown in fixed scaling; (2) Min-max scaling effectively addresses the scoring bias by spreading its predictions across the target score range; (3) Outlier clipping further alleviates the distortion of the distribution caused by the outliers, contributing to robust performance.

\section{Related Work}
\paragraph{Automated Essay Scoring.} Discovering essay representation discriminative of writing qualities has been a major concern in AES. Early works had explored extracting hand-crafted linguistic features \citep{yannakoudakis-etal-2011-new, persing-ng-2013-modeling, chen-he-2013-automated}, learning features via neural networks \citep{taghipour-ng-2016-neural, dong-etal-2017-attention, tay2018skipflow}, as well as combining the two \citep{dasgupta-etal-2018-augmenting, uto-etal-2020-neural}. Particularly, recent works \citep{yang-etal-2020-enhancing, xie-etal-2022-automated} exhibited high level of agreement with human raters in the prompt-specific setting.

Despite the success of prompt-specific models, they experienced significant performance drops when adapted to unseen prompts \citep{jin-etal-2018-tdnn, cozma-etal-2018-automated}, limiting their applicability in practice. To tackle this challenge, recent works 
introduce domain adaptation \citep{phandi2015flexible, cao2020domain} and generalization \citep{ridley2020prompt, jiang-etal-2023-improving} techniques to AES. Nevertheless, these methods still consume considerable amount of labelled essays of the source domain(s).

The advent of LLM and its versatility on a wide range of downstream tasks \citep{wei2022chain, kojima2022large} has raised attention to its potential in essay scoring, giving rise to LLM-based zero-shot AES. For instance, \citet{mizumoto2023exploring} prompt text-davinci-003 to only respond with the essay's overall score referring to a given scoring rubric. \citet{yancey2023rating} reveal that LLMs benefit from referring to a detailed rubric or generating a rationale prior to the score in zero-shot setting. However, GPT-4 with these strategies still exhibits suboptimal performance similar to length-only classifier which relies solely on essay's character length for its prediction.

\paragraph{LLM-based Evaluators.} Recent studies have investigated the use of LLM as a reference-free evaluation metric for natural language generation (NLG) tasks \citep{fu2023gptscore, chiang-lee-2023-large}. \citet{zheng2023judging} showed that strong LLMs (e.g., GPT-4) highly matched human preferences when evaluating the quality of AI assistant's respones in open-ended QA tasks. Furthermore, LLMs achieved state-of-the-art agreement with human judgements in translation quality assessment \citep{kocmi-federmann-2023-large} and summarization task \citep{wang-etal-2023-chatgpt}, verifying LLM evaluators' scalability to a variety of NLG tasks.

In addition to the variations in prompt design to fit different tasks, sophisticated prompting strategies have been devised to further elicit LLMs' potential as evaluators: \citet{liu-etal-2023-g} prompt LLM with auto-generated evaluation steps to elicit CoT, achieving superior correlation with human ratings on text summarization and dialogue generation task; \citet{chan2023chateval, zhang2023wider} devised a multi-agent cooperation framework where LLMs with diverse role descriptions synergize to refine the evaluation result. 


\section{Conclusion}
We present MTS (Multi Trait Specialization), an LLM-based zero-shot prompting framework for AES. In essence, MTS leverages trait-specific conversations with LLM to derive the overall score from diverse aspects of writing proficiency. Experimental results show that MTS consistently outperforms Vanilla approach in average QWK across all LLMs and datasets, while also substantially reducing the gap with fully supervised SOTA model. Our analysis reveals key insights into the success of MTS: (1) Providing the LLM with predefined scoring criteria regulates its scoring behavior, contributing to improved performance; (2) MTS benefits from subproblem decomposition such as independent trait-specific conversations and separation of quote retrieval task and scoring task; (3) Outlier clipping and min-max scaling effectively map the predictions to arbitrary target score range, alleviating the LLM's scoring bias as well as min-max scaling's sensitivity to the outliers.  

\section*{Acknowledgments}
This work is supported by the National Natural Science Foundation of China (62076008) and the Key Project of Natural Science Foundation of China (61936012).

\section*{Limitations}

First, MTS consumes significantly longer inference time than Vanilla approach due to multiple rounds of conversation. In addition, the pre-generated scoring criteria is included in the conversations for every essay, further increasing the computational cost. We believe that distilling the predictions to small models such as BERT \citep{devlin-etal-2019-bert} would be a promising direction for cost-effective inference. Second, our analysis still demands more detailed illustrations of the LLMs' scoring behavior. For instance, LLMs produce verbal evaluations on the quotes they found in the essay, but it is unclear whether there is a faithful relation between the evaluations and the score. Moreover, it should be examined if the inclusion of scoring criteria truly leads to a more consistent scoring behavior, or it merely shifts the average of predicted scores closer to the groundtruth score while maintaining the same degree of inconsistency. Third, outlier clipping using Q1 and Q3 may not make min-max scaling completely resistant to outliers. We have not conducted extensive experiments with more robust ways of addressing outliers.


\section*{Ethics Statement}
\paragraph{Potential Risks}
Our method does not guarantee fair evaluations, that is, MTS might reinforce LLMs' scoring tendency of favoring certain social groups. For example, it is possible that the predictive outcomes assign higher scores to a group with certain L1 (first language) background than the others. In addition, the datasets (ASAP and TOEFL11) we use might disproportionately represent certain demographic group, potentially leading to a biased conclusion. We partially address this concern by selecting TOEFL11 test dataset that contains equal number of samples from each L1. For ASAP, there is no demographic information provided.

\paragraph{Use of Scientific Artifacts}
We use the open source scikit-learn package (v1.0.2) \citep{scikit-learn} for the calculation of QWK. We conduct experiments with ASAP \citep{asap-aes} and TOEFL11 \citep{blanchard2013toefl11} datasets, which are available for non-commercial research purposes. ASAP have anonymized personally identifying information from the essays by replacing them with symbols. TOEFL11 only includes essays with the author's permission for research use. As for the LLMs used in our study, OpenAI authorizes exploring its LLMs including ChatGPT \cite{ChatGPT} through its API for research publication (see OpenAI's sharing and publication policy). Llama 2 \citep{touvron2023llama} and Mistral 7b \cite{jiang2023mistral} are licensed under the Llama 2 Community license and Apache-2.0 license, respectively, both permitting research use.

\paragraph{Computational Budget}
We use a single NVIDIA A40 for each model inference including Llama2-7b-chat, Llama2-13b-chat and Mistral-7b-Instruct-v0.2. With only one sample in each batch, running MTS on ASAP test set takes approximately one GPU day for Llama2-7b-chat and Mistral-7b-Instruct-v0.2 and two GPU days for Llama2-13b-chat. Running time of MTS for ChatGPT through OpenAI API was similar to that of Llama2-13b-chat.



\bibliography{custom}

\appendix


\section{Instructions for Multi Trait Decomposition} \label{appendix:mtd}
For multi trait decomposition, we prompt ChatGPT with different instructions for ASAP and TOEFL11. While ASAP's rubric guidelines do not necessarily consist of four traits, TOEFL11's rubric guideline (here we use IELTS Task2 Writing Band Descriptor, see Section~\ref{sec:implementation_details}) explicitly divides writing proficiency into four traits. Therefore, for ASAP, we ask ChatGPT to first determine the four traits and generate scoring criteria for each trait. In contrast, for TOEFL11, we ask it to generate scoring criteria based on the traits determined by the rubric guideline. Specifically, we use the following instructions for ASAP and TOEFL11 where the contents to be filled are substituted with comments between double curly braces {\fontfamily{cmtt}\selectfont\{\{ \}\}}: \newline
\noindent\textbf{ASAP} \newline
{\fontfamily{cmtt}\selectfont
[Excerpt] \newline
\{\{excerpt (specific to prompt3-6)\}\} \newline
(end of [Excerpt]) \newline
[Prompt] \newline
\{\{prompt\}\} \newline
(end of [Prompt]) \newline
[Rubric Guidelines] \newline
\{\{rubric guidelines\}\} \newline
(end of [Rubric Guidelines]) \newline
Refer to the provided [Prompt] and [Rubric Guidelines] to generate an essay scoring rubric divided into four primary dimensions of writing quality. Adhere to the requirements of [Prompt] and [Rubric Guidelines] when you determine the four dimensions of writing quality. At each dimension, make sure a brief description of the dimension is added before the scoring criteria. The score scale of each dimension ranges from 0 to 10, and the total score is 40. \newline
}

\noindent\textbf{TOEFL11} \newline
{\fontfamily{cmtt}\selectfont
[Scoring Rubric] \newline
\{\{IELTS Task2 Writing Band Descriptor, restricted to one trait\}\} \newline
(end of [Scoring Rubric]) \newline
Refer to [Scoring Rubric] to generate a scoring criteria with score ranging from 0 to 10, following  the instruction below: \newline
1. Briefly describe “\{\{trait\}\}” with one sentence. \newline
2. Divide the score range [0-10] into 5 appropriate intervals. \newline
3. For each interval, summarize its characteristics. \newline
}

\section{Guidance from Multi Trait Decomposition} \label{appendix:guidance_mtd}
We present guidance from multi-trait decomposition, comprising: (1) trait, (2) trait description, and (3) scoring criteria. MTS uses (1) and (2) for the role prompt (system message) and (1) and (3) for conversations. ASAP uses different rubric guidelines for each of its eight prompts, generating distinct guidance for each, while TOEFL11 applies the same rubric for all eight, resulting in identical guidance. Due to length, we show only the guidance for ASAP Prompt 1 and all prompts for TOEFL11.\newline

\noindent\textbf{ASAP Prompt 1}\newline
\textit{
Position and Thesis Clarity\newline
This dimension evaluates how clearly the writer takes a stance on the effects of computers on people and how effectively this stance is conveyed in the thesis statement.\newline
*Scoring Criteria:*\newline
- 0-2: The position is unclear or absent. The thesis lacks a clear stance or is entirely missing.\newline
- 3-5: The position is somewhat evident but lacks clarity or specificity in the thesis statement.\newline
- 6-8: The position is clear, though it may require further specificity or nuance in the thesis statement.\newline
- 9-10: The position is crystal clear, and the thesis statement effectively communicates the writer's stance with precision and depth.\newline \newline
Supporting Details and Evidence \newline
This dimension assesses the quality and relevance of the supporting details and evidence used to back the writer's position. \newline
*Scoring Criteria:* \newline
- 0-2: Very minimal or no supporting details provided.\newline
- 3-5: General or vague supporting details with minimal relevance to the thesis.\newline
- 6-8: Adequate supporting details offered, although some lack specificity or relevance.\newline
- 9-10: Rich and specific supporting details effectively back the thesis, providing compelling evidence and relevance.\newline \newline
Organization and Structure\newline
This dimension evaluates the overall coherence, logical progression, and structural framework of the essay.\newline
*Scoring Criteria:*\newline
- 0-2: The essay lacks organization and structure, making it challenging to follow or understand.
- 3-5: Shows minimal organization but lacks a coherent structure or transitions.\newline
- 6-8: Demonstrates satisfactory organization with some coherence, though transitions may be weak in connecting ideas.\newline
- 9-10: Strong organization with clear and smooth transitions, presenting ideas logically and coherently.\newline \newline
Style, Language, and Audience Awareness\newline
This dimension assesses the writer's language use, style, and their ability to engage the audience while demonstrating an awareness of the target readers.\newline
*Scoring Criteria:*\newline
- 0-2: Language use is awkward, and there's no evident awareness of the audience.\newline
- 3-5: Language use is basic, and there's little attempt to engage the audience or demonstrate awareness.\newline
- 6-8: Language is somewhat engaging, with occasional attempts to connect with the audience.\newline
- 9-10: Language is engaging, sophisticated, and consistently demonstrates an acute awareness of the audience, effectively connecting with them.\newline
}

\noindent\textbf{TOEFL11} \newline
\textit{
Task Response\newline
This dimension evaluates how well the prompt is understood, addressed, and developed within the response.\newline
0-2:\newline
- Barely relevant or unrelated content to the given prompt.\newline
- Lack of identifiable position or comprehension of the question.\newline
- Minimal or no development of ideas; content may be tangential or copied.\newline
3-4:\newline
- Partially addresses the prompt but lacks depth or coherence.\newline
- Discernible position, but unclear or lacking in support.\newline
- Ideas are difficult to identify or irrelevant with some repetition.\newline
5-6:\newline
- Addresses main parts of the prompt but incompletely or with limited development.\newline
- Presents a position with unclear or repetitive development.\newline
- Some relevant ideas but insufficiently developed or supported.\newline
7-8:\newline
- Adequately addresses the prompt with clear and developed points.\newline
- Presents a coherent position with well-extended and supported ideas.\newline
- Some tendencies toward over-generalization or lapses in content, but mostly on point.\newline
9-10:\newline
- Fully and deeply explores the prompt with a clear, well-developed position.\newline
- Extensively supported ideas relevant to the prompt.\newline
- Extremely rare lapses in content or support; demonstrates exceptional depth and insight.\newline \newline
Coherence and Cohesion\newline
This criterion assesses how well ideas are logically organized and connected within a written response.\newline
0-2:\newline
- Lack of coherence; response is off-topic or lacking in relevant message.\newline
- Minimal evidence of organizational control or logical progression.\newline
- Virtually absent or ineffective use of cohesive devices and paragraphing.\newline
3-4:\newline
- Ideas are discernible but arranged incoherently or lack clear progression.\newline
- Unclear relationships between ideas, limited use of basic cohesive devices.\newline
- Minimal or unclear referencing, inadequate paragraphing if attempted.\newline
5-6:\newline
- Some underlying coherence but lacks full logical organization.\newline
- Relationships between ideas are somewhat clear but not consistently linked.\newline
- Limited use of cohesive devices, with inaccuracies or overuse, and occasional repetition.\newline
- Inconsistent or inadequate paragraphing.\newline
7-8:\newline
- Generally organized with a clear overall progression of ideas.\newline
- Cohesive devices used well with occasional minor lapses.\newline
- Effective paragraphing supporting coherence, though some issues in sequencing or clarity within paragraphs.\newline
9-10:\newline
- Effortless follow-through of ideas with superb coherence.\newline
- Seamless and effective use of cohesive devices with minimal to no lapses.\newline
- Skilful paragraphing enhancing overall coherence and logical progression.\newline \newline
Lexical Resource\newline
This dimension evaluates the range, precision, and appropriateness of vocabulary used within a written response.\newline
0-2:\newline
- Minimal to no resource evident; extremely limited vocabulary or reliance on memorized phrases.\newline
- Lack of control in word formation, spelling, and recognition of vocabulary.\newline
- Communication severely impeded due to the absence of lexical range.\newline
3-4:\newline
- Inadequate or limited resource; vocabulary may be basic or unrelated to the task.\newline
- Possible dependence on input material or memorized language.\newline
- Errors in word choice, formation, or spelling impede meaning.\newline
5-6:\newline
- Adequate but restricted resource for the task.\newline
- Limited variety and precision in vocabulary, causing simplifications and repetitions.\newline
- Noticeable errors in spelling/word formation, with some impact on clarity.\newline
7-8:\newline
- Sufficient resource allowing flexibility and precision in expression.\newline
- Ability to use less common or idiomatic items, despite occasional inaccuracies.\newline
- Some errors in spelling/word formation with minimal impact on communication.\newline
9-10:\newline
- Full flexibility and precise use of a wide range of vocabulary.\newline
- Very natural and sophisticated control of lexical features with rare minor errors.\newline
- Skilful use of uncommon or idiomatic items, enhancing overall expression.\newline \newline
Grammatical Range and Accuracy\newline
This dimension assesses the breadth of grammatical structures used and the precision in applying them within written communication.\newline
0-2:\newline
- Absence or extremely limited evidence of coherent sentence structures.\newline
- Lack of control in grammar, minimal to no use of sentence forms.\newline
- Language largely incomprehensible or irrelevant to the task.\newline
3-4:\newline
- Attempts at sentence forms but predominantly error-laden.\newline
- Inadequate range of structures with frequent grammatical errors.\newline
- Limited coherence due to significant errors impacting meaning.\newline
5-6:\newline
- Limited variety in structures; attempts at complexity with faults.\newline
- Some accurate structures but with noticeable errors and repetitions.\newline
- Clear attempts at complexity but lacking precision and fluency.\newline
7-8:\newline
- Adequate variety with some flexibility in using complex structures.\newline
- Generally well-controlled grammar but occasional errors.\newline
- Clear attempts at complexity and flexibility in sentence structures.\newline
9-10:\newline
- Extensive range with full flexibility and precision in structures.\newline
- Virtually error-free grammar and punctuation.\newline
- Exceptional command with rare minor errors, showcasing nuanced and sophisticated language use.
}

\section{Prompt Templates}\label{appendix:template}
In this section, we provide the exact templates of the prompts used for Vanilla and MTS. Our prompt design consists of three components: system message, user message and assistant message. Contents to be filled are placed between double curly braces {\fontfamily{cmtt}\selectfont\{\{ \}\}}. Contents specific to ASAP are enclosed with {\fontfamily{cmtt}\selectfont \textit{ASAP($\cdot$)}} and those specific to TOEFL11 are enclosed with {\fontfamily{cmtt}\selectfont \textit{TOEFL11($\cdot$)}}, both in italic font.

\subsection{Template for MTS}\label{appendix:template_mts}

\noindent\textbf{System Message} \newline
{\fontfamily{cmtt}\selectfont
You are a member of the English essay writing test evaluation committee. Four teachers will be provided with a [Prompt] and an [Essay] written by a student in response to the [Prompt]. Each teacher will score the essays based on different dimensions of writing quality. Your specific responsibility is to score the essays in terms of “\{\{trait\}\}". \{\{a brief description of the trait\}\} Focus on the content of the [Essay] and the [Scoring Rubric] to determine the score. \newline
}

\noindent\textbf{User Message} \newline
{\fontfamily{cmtt}\selectfont
[Prompt] \newline
\{\{prompt\}\} \newline
(end of [Prompt]) \newline
\textit{ASAP([Note] \newline
I have made an effort to remove personally identifying information from the essays using the Named Entity Recognizer (NER). The relevant entities are identified in the text and then replaced with a string such as "\{PERSON\}", "\{ORGANIZATION\}", "\{LOCATION\}", "\{DATE\}", "\{TIME\}", "\{MONEY\}", "\{PERCENT\}”, “\{CAPS\}” (any capitalized word) and “\{NUM\}” (any digits)\footnote{In the original ASAP dataset, all named entities are marked in the format of "@named entity". We convert this format to "\{named entity\}" in order to make the boundaries of the named entities more explicit.}. Please do not penalize the essay because of the anonymizations. \newline
(end of [Note]))}\newline
[Essay] \newline
\{\{essay\}\} \newline
(end of [Essay]) \newline
Q. List the quotations from the [Essay] that are relevant to “\{\{trait\}\}” and evaluate whether each quotation is well-written or not. \newline
}

\noindent\textbf{Assistant Message} \newline
{\fontfamily{cmtt}\selectfont
\{\{a response from the LLM\}\} \newline
}

\noindent\textbf{User Message} \newline
{\fontfamily{cmtt}\selectfont
[Scoring Rubric] \newline
**\{\{trait\}\}**: \newline
\{\{scoring criteria\}\} \newline
(end of [Scoring Rubric]) \newline
Q. Based on the [Scoring Rubric] and the quotations you found, how would you rate the “\{\{trait\}\}” of this essay? 
Assign a score from 0 to 10, strictly following the [Output Format] below. \newline
[Output Format] \newline
Score: <score>insert ONLY the numeric score (from 0 to 10) here</score> \newline
(End of [Output Format])
}

\subsection{Template for Vanilla}\label{appendix:template_vanilla}
\noindent\textbf{System Message} \newline
{\fontfamily{cmtt}\selectfont
As an English teacher, your primary responsibility is to evaluate the writing quality of essays written by \textit{ASAP(middle school students)} \textit{TOEFL11(second language learners on an English exam)}. During the assessment process, you will be provided with a prompt and an essay. First, you should provide comprehensive and conrete feedback that is closely linked to the content of the essay. It is essential to avoid offering generic remarks that could be applied to any piece of writing. To create a compelling evaluation for both the student and fellow experts, you should reference specific content of the essay to substantiate your assessment. Next, your evaluation should culminate in assigning an overall score to the student's essay, \textit{ASAP(measured on a scale from \{\{minimum score value\}\} to \{\{maximum score value\}\}, where higher score should reflect a higher level of writing quality. It's crucial to tailor your evaluation criteria to be well-suited for middle school level writing, taking into account the developmental stage and capabilities of these students.)} \textit{TOEFL11(on a three level scale of "low", "medium" and "high". It's crucial to tailor your evaluation criteria to be well-suited for second language learners, taking into account their expected abilities.)} \newline
}

\noindent\textbf{User Message} \newline
{\fontfamily{cmtt}\selectfont
[Prompt] \newline
\{\{prompt\}\} \newline
(end of [Prompt]) \newline
\textit{ASAP([Note] \newline
I have made an effort to remove personally identifying information from the essays using the Named Entity Recognizer (NER). The relevant entities are identified in the text and then replaced with a string such as "\{PERSON\}", "\{ORGANIZATION\}", "\{LOCATION\}", "\{DATE\}", "\{TIME\}", "\{MONEY\}", "\{PERCENT\}”, “\{CAPS\}” (any capitalized word) and “\{NUM\}” (any digits). Please do not penalize the essay because of the anonymizations. \newline
(end of [Note]))} \newline
[Essay] \newline
\{\{essay\}\} \newline
(end of [Essay]) \newline
Strictly follow the format below to give your answer. Other formats are NOT allowed. 
Evaluation: <evaluation>insert evaluation here</evaluation> \newline
\textit{ASAP(Score: <score>insert score (\{\{minimum score value\}\} to \{\{maximum score value\}\}) here</score>)} \textit{TOEFL11(Score: <score>insert score (choose one of "low", "medium", and "high") here</score>)}
}

\subsection{Template for Scoring All Traits Sequentially} \label{appendix:template_sequential}
\noindent\textbf{System Message} \newline
{\fontfamily{cmtt}\selectfont
You are an English teacher who is responsible for rating essays. You will be provided with a prompt and a student’s essay written in response to the prompt. Follow the provided [Evaluation Steps] and assign a score to the essay in the specified format. \newline 
}

\noindent\textbf{User Message} \newline
{\fontfamily{cmtt}\selectfont
[Prompt] \newline
\{\{prompt\}\} \newline
(end of [Prompt]) \newline
\textit{ASAP([Note] \newline
I have made an effort to remove personally identifying information from the essays using the Named Entity Recognizer (NER). The relevant entities are identified in the text and then replaced with a string such as "\{PERSON\}", "\{ORGANIZATION\}", "\{LOCATION\}", "\{DATE\}", "\{TIME\}", "\{MONEY\}", "\{PERCENT\}”, “\{CAPS\}” (any capitalized word) and “\{NUM\}” (any digits). Please do not penalize the essay because of the anonymizations. \newline
(end of [Note]))} \newline
[Essay] \newline
\{\{essay\}\} \newline
(end of [Essay]) \newline
[Evaluation Steps] \newline
\{\{evaluation steps which include the entire guidance generated from multi trait decomposition (see Appendix~\ref{appendix:guidance_mtd})\}\} \newline
(end of [Evaluation Steps]) \newline
Q. For each step in [Evaluation Steps], assign a score from 0 to 10, strictly following the [Output Format] below. \newline
[Output Format] \newline
Step 1 \newline
- Evaluation: <evaluation>insert evaluation here</evaluation> \newline
- Score: <score>insert ONLY the numeric score (from 0 to 10) here</score> \newline
Step 2 \newline
- Evaluation: <evaluation>insert evaluation here</evaluation> \newline
- Score: <score>insert ONLY the numeric score (from 0 to 10) here</score> \newline
Step 3 \newline
- Evaluation: <evaluation>insert evaluation here</evaluation> \newline
- Score: <score>insert ONLY the numeric score (from 0 to 10) here</score> \newline
Step 4 \newline
- Evaluation: <evaluation>insert evaluation here</evaluation> \newline
- Score: <score>insert ONLY the numeric score (from 0 to 10) here</score> \newline
(end of [Output Format]) \newline
}

\subsection{Template for Scoring Each Trait Independently} \label{appendix:template_independent}
\noindent\textbf{System Message} \newline
{\fontfamily{cmtt}\selectfont
You are a member of the English essay writing test evaluation committee. Four teachers will be provided with a [Prompt] and an [Essay] written by a student in response to the [Prompt]. Each teacher will score the essays based on different dimensions of writing quality. Your specific responsibility is to score the essays in terms of “\{\{trait\}\}". \{\{a brief description of the trait\}\} Focus on the content of the [Essay] and the [Scoring Rubric] to determine the score. \newline
}

\noindent\textbf{User Message} \newline
{\fontfamily{cmtt}\selectfont
[Prompt] \newline
\{\{prompt\}\} \newline
(end of [Prompt]) \newline
\textit{ASAP([Note] \newline
I have made an effort to remove personally identifying information from the essays using the Named Entity Recognizer (NER). The relevant entities are identified in the text and then replaced with a string such as "\{PERSON\}", "\{ORGANIZATION\}", "\{LOCATION\}", "\{DATE\}", "\{TIME\}", "\{MONEY\}", "\{PERCENT\}”, “\{CAPS\}” (any capitalized word) and “\{NUM\}” (any digits). Please do not penalize the essay because of the anonymizations. \newline
(end of [Note]))}\newline
[Essay] \newline
\{\{essay\}\} \newline
(end of [Essay]) \newline
[Scoring Rubric] \newline
**\{\{trait\}\}**: \newline
\{\{scoring criteria\}\} \newline
(end of [Scoring Rubric]) \newline
Q: From the above [Scoring Rubric], how would you rate the “\{\{trait\}\}” of this essay? Respond a reasoning followed by a score from 0 to 10, strictly following the [Output Format] below: \newline
[Output Format] \newline
Reasoning: <reasoning>insert your reasoning which will justify your decision on the score</reasoning> \newline
Score: <score>insert ONLY the numeric score (from 0 to 10) here</score>
(End of [Output Format]) \newline
}

\section{Details of Re-implementation of NPCR} \label{appendix:details_npcr}

NPCR \cite{xie-etal-2022-automated} was originally implemented on ASAP. For our re-implementation on ASAP, while we leave out the same test set as MTS, the remaining data is randomly divided into train set and validation set by $4:1$. We re-implement NPCR on TOEFL11 as well with minimal adjustments: we use the train, dev and test split provided by \citet{tetreault2013report}. The test set is identical to that of MTS. The predicted scores are scaled to $[1,5]$ and mapped to low/medium/high with respect to the thresholds $[2.25, 3.75]$, which is consistent with Section \ref{sec:implementation_details}. 

For both datasets, the number of epochs is reduced from $80$ to $20$ so that the time for training and inference is kept in an acceptable range.  Other settings are identical to the original implementation.

\end{document}